\documentclass{article}

\usepackage[preprint]{neurips_2021}

\usepackage[utf8]{inputenc} 
\usepackage[T1]{fontenc}    
\usepackage{hyperref}       
\usepackage{url}            
\usepackage{booktabs}       
\usepackage{amsfonts}       
\usepackage{nicefrac}       
\usepackage{microtype}      
\usepackage{xcolor}         
\usepackage[pdftex]{graphicx}
\usepackage{algorithm}

\title{Dropout against Deep Leakage from Gradients}

\author{%
  Yanchong Zheng \\
  School of Computer Science\\
  Wuhan University\\
  \texttt{gazlayxyc@whu.edu.cn} \\
}

\begin{document}

\maketitle

\begin{abstract}

As the scale and size of the data increases significantly nowadays, federal learning (\citet{bonawitz2019federated}) for high performance computing and machine learning has been much more important than ever before (\citet{abadi2016tensorflow}). People used to believe that sharing gradients seems to be safe to conceal the local training data during the training stage. However, \citet{zhu19deep} demonstrated that it was possible to recover raw data from the model training data by detecting gradients. They use generated random dummy data and minimise the distance between them and real data. \citet{zhao2020idlg} pushes the convergence algorithm even further. By replacing the original loss function with cross entropy loss, they achieve better fidelity threshold. In this paper, we propose using an additional dropout (\citet{srivastava2014dropout}) layer before feeding the data to the classifier. It is very effective in preventing leakage of raw data, as the training data cannot converge to a small RMSE even after 5,800 epochs with dropout rate set to 0.5.

\end{abstract}

\section{Introduction}

The increasing demand for accuracy has lead to larger model sizes than ever before. Therefore, distributed training and federal learning is see their increasing popularity (\citet{Barney2009}, \citet{NIPS2012_6aca9700}). Distributed computing on cloud clusters makes it possible for users to train extremely large models while maintaining a reasonable training time. However, in a typical map-reduce setting, there will be multiple processes of splitting the data and sending them to different computers. This scheme is named federal learning, which is widely adapted on private datasets. For example, cellphone manufacturing companies use users' activity logs to predict their habits and favorite applications.

For a long time, people used to believe that medium training data was safe to share across different computers. Is the medium training data able to reveal any information about the original local data? \citet{zhu19deep} found that it was possible to recover original data from the medium model training data. In \citet{zhu19deep} work, they propose using the following methods to recover raw data from training data. Firstly, they generate a pair of dummy inputs and labels randomly and feed them into the model, performing the standard forward and backward. Secondly, instead of updating model weights in a typical setting, they use the inputs and labels to minimize the distance between fake gradients and actual gradients, trying to use the gradients to match fake data to original ones. If the optimization converges to a small number in the end, the private training data is eventually recovered. \citet{zhao2020idlg} propose a simple and valid approach to recover the real labels from the shared gradients. They use cross entropy as the loss function and find out that the gradient of the classification loss the correct label activation. What's more, their method also reaches a better fidelity threshold than the original experiment.

In this work, we propose a simple yet effective approach: we use dropout (\citet{srivastava2014dropout}) to increase the noise and randomness of the data. We add an additional dropout layer between the encoder and the classifier, and evaluate different settings of dropout rate to explore the mechanism behind the phenomenon. With carefully crafted settings of the dropout rate, even the simplest function can achieve a relatively great result.

\citet{hscla20} aims to solve the challenge of mitigating privacy risk in collaborative learning without slowing down training or reducing accuracy. TextHide requires the participating computers to add an encryption step in order to prevent eavesdropping attacks. \citet{datamix20} proposes a similar method. Its DataMix framework aims to mediate between limited computation resource and privacy sensitive cloud servers by mixing and de-mixing the data.

As deep leakage posed a non-negligible threat to federal learning, using a dropout layer allows the researchers and engineers to prevent such attacks with minimal efforts. What's more, as dropout has been a common technique in transformers from natural language processing, we believe it would be a prevalent method in preserving user privacy.

Our contributions include:
\begin{itemize}
    \item we demonstrate that it is possible to defend deep leakage from gradients using a simple dropout layer between the encoder model and classifier.
    \item we explore the mechanisms of dropout in preventing deep leakage attack.
\end{itemize}

\section{Related Work}

\subsection{Deep Leakage from Gradients}

In the past, shallow leakages required extra information and could only generate similar images. \citet{zhu19deep} discovers a much stronger attack approach than previous methods, which is able to recover data pixel-wise for images and token-wise for texts. The deep leakage has become a severe challenge to multi-node distributed learning clusters, as the fundamental gradient sharing scheme is not reliable anymore to protect the privacy of the user. \citet{zhao2020idlg} propose an improved method. They use cross entropy loss instead of the original loss function, and their experiments result in better fidelity threshold and accuracy in extracting correct labels from the random generated dummy data.

\subsection{Methods against Deep Leakage}

Data privacy has become a major concern in deep learning, especially cloud machine learning, where gradients are shared across different machines. \citet{hscla20} designed TextHide for plugging into popular machine learning framework, which encrypts the output vectors from the encoder by mixing data with neighboring items. \citet{datamix20} introduces a privacy-preserving edge-cloud inference framework, which is called DataMix. In order to take advantage of abundant computing resources on the cloud and maintain privacy in the meantime, they also mix and de-mix the data before transferring.

\subsection{Dropout}

Although large neural networks are extremely powerful at producing reliable predictions, over-fitting becomes an unavoidable problem for such networks. \citet{srivastava2014dropout} proposed dropout, and its main idea is to randomly drop data during the training process. By setting some data from exponential numbers to zero, the effect of dropout is very similar to ensemble models producing an average output. It has shown overwhelming effects in reducing over-fitting and improves the performance in various tasks, such as image classification, speech recognition, natural language processing, etc.

\section{Method}

\begin{figure}
    \centering
    \includegraphics[width=0.75\linewidth]{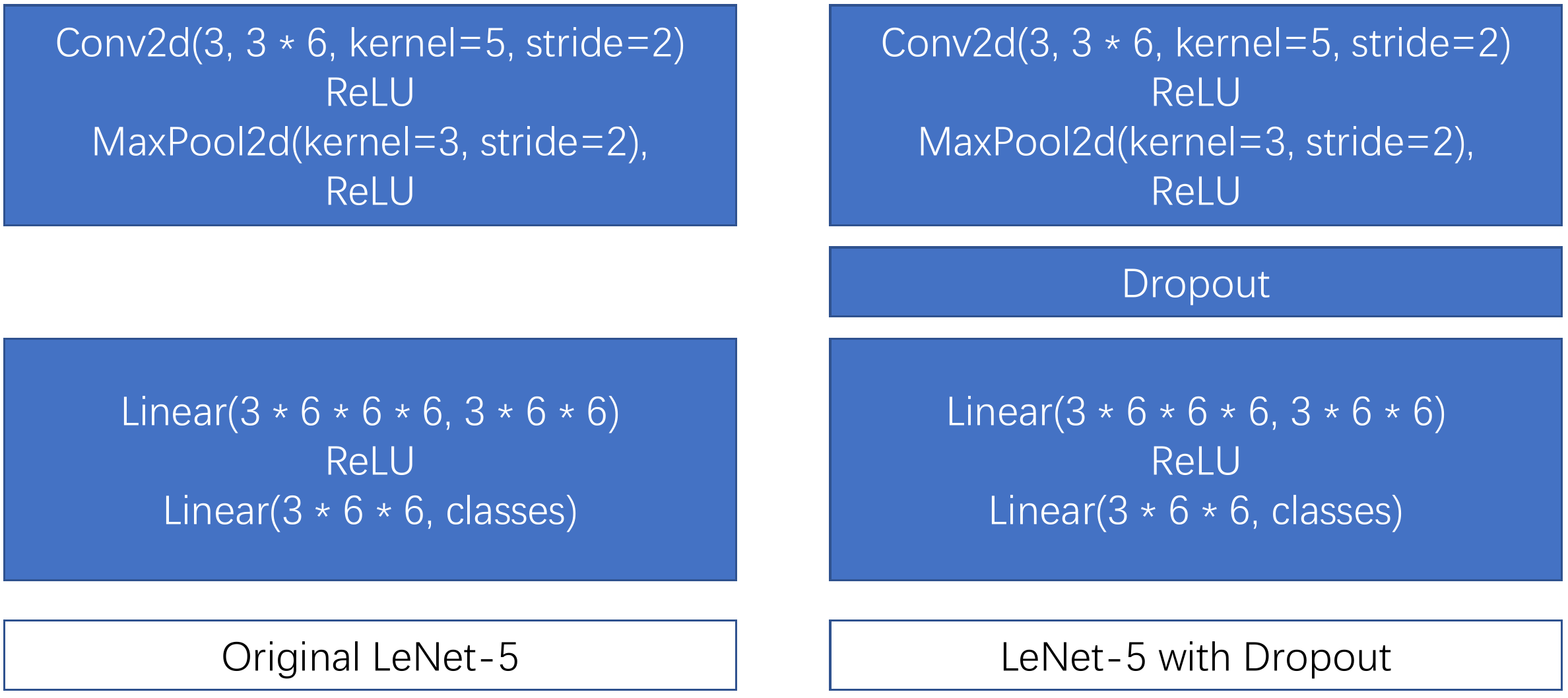}
    \caption{Original Model and Model with Dropout}
    \label{fig:model}
\end{figure}

We show that stealing an image pixel-wise can be prevented by adding an additional dropout layer between the encoder and the classifier, as it is illustrated in Figure \ref{fig:model}.

\begin{figure}
    \centering
    \includegraphics[width=0.5\linewidth]{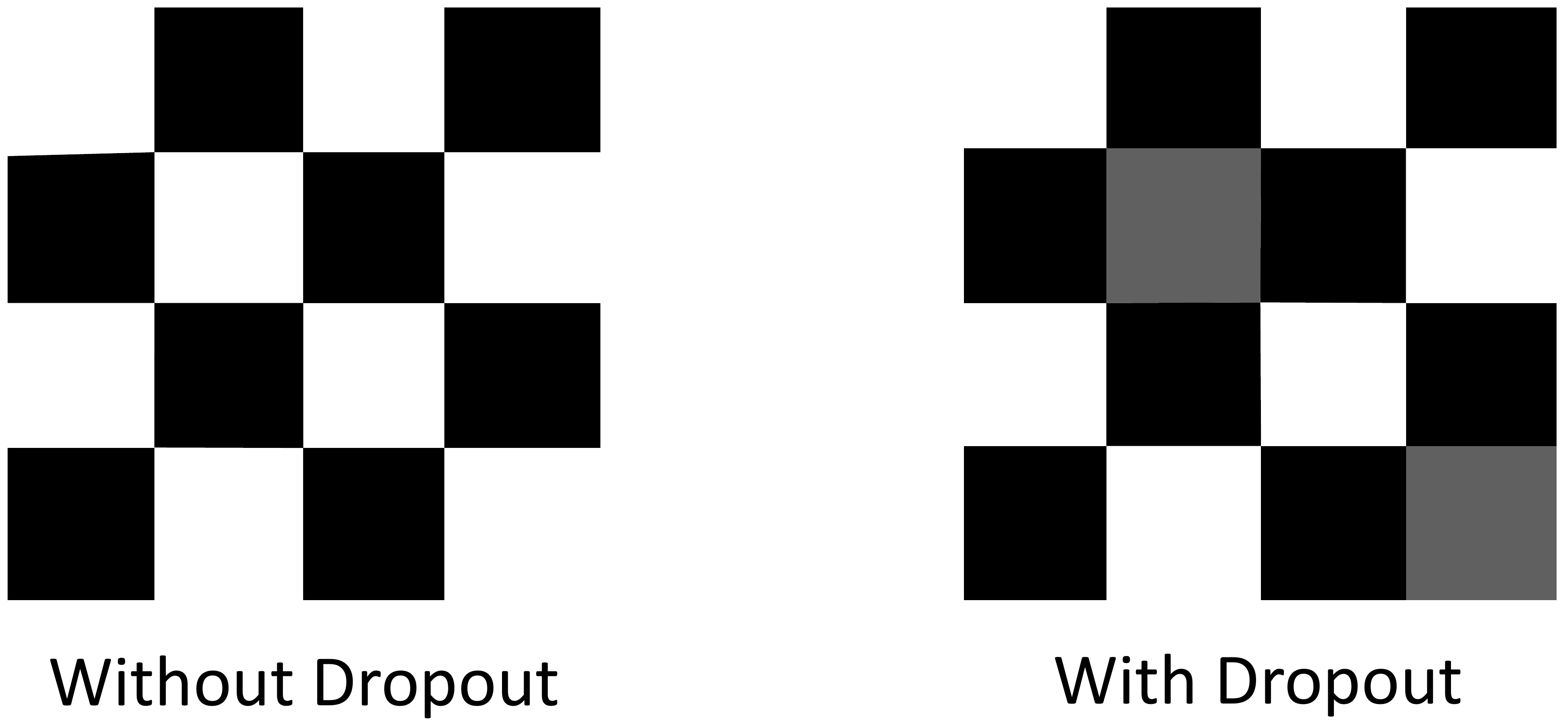}
    \caption{Black and White for Original Data, Grey for Zero}
    \label{fig:mechanism}
\end{figure}

As illustrated in Figure \ref{fig:mechanism}, dropout makes some data become zero. Therefore, dropout makes some of the data irrelevant to the local private data. The converge algorithm has more directions, which slows down the convergence process. When the number of directions becomes big enough, the possibility becomes small enough and the model becomes much safer.

\section{Experiment}

\subsection{Experiment Setup}

\paragraph{Dataset} We run the experiment on CIFAR-10, which is a dataset collected by \citet{cifar10}. CIFAR-10 has 60,000 color images of 32*32 resolution in 10 classes, and each class has 6,000 images. For each image, we run 5,600 epochs.

\paragraph{Model} In the experiment, we use the LeNet-5 Model as the experiments in Deep Leakage from Gradients. It is a simple convolutional neural network proposed by \citet{726791}. The only modification we do is adding a dropout layer between the encoder and the classifier. 

\paragraph{Converge Algorithm} We use the same algorithm as Improved Deep Leakage from Gradients. Firstly, the algorithm extracts the real label. Secondly, it initializes dummy data, which is image data and label data respectively. Thirdly, it calculates the loss and minimises it. Finally, it updates the dummy data to match the gradients. To be more general, all the weights are randomly initiated. The learning rate is set at 1.0.

\subsection{Experiment Result}

\begin{figure}
    \centering
    \includegraphics[width=1\linewidth]{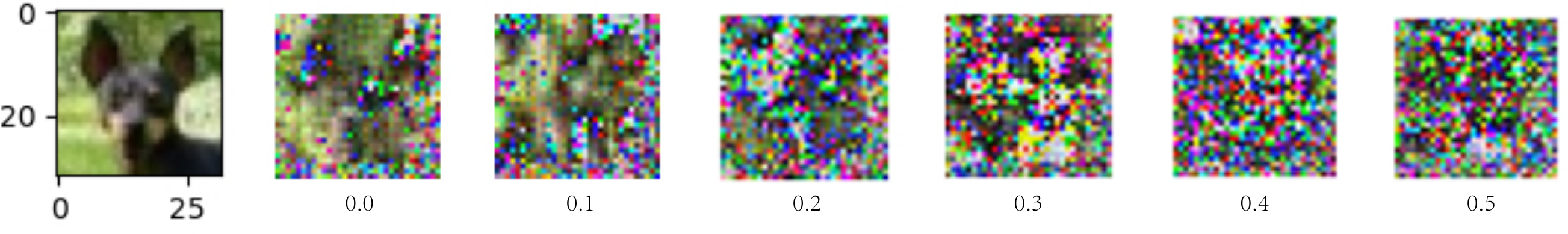}
    \caption{A randomly selected sample from the result set, from left to right is Ground Truth, Dropout Rate from 0.0 to 0.5}
    \label{fig:0.0-0.5}
\end{figure}

\begin{figure}
    \centering
    \includegraphics[width=0.5\linewidth]{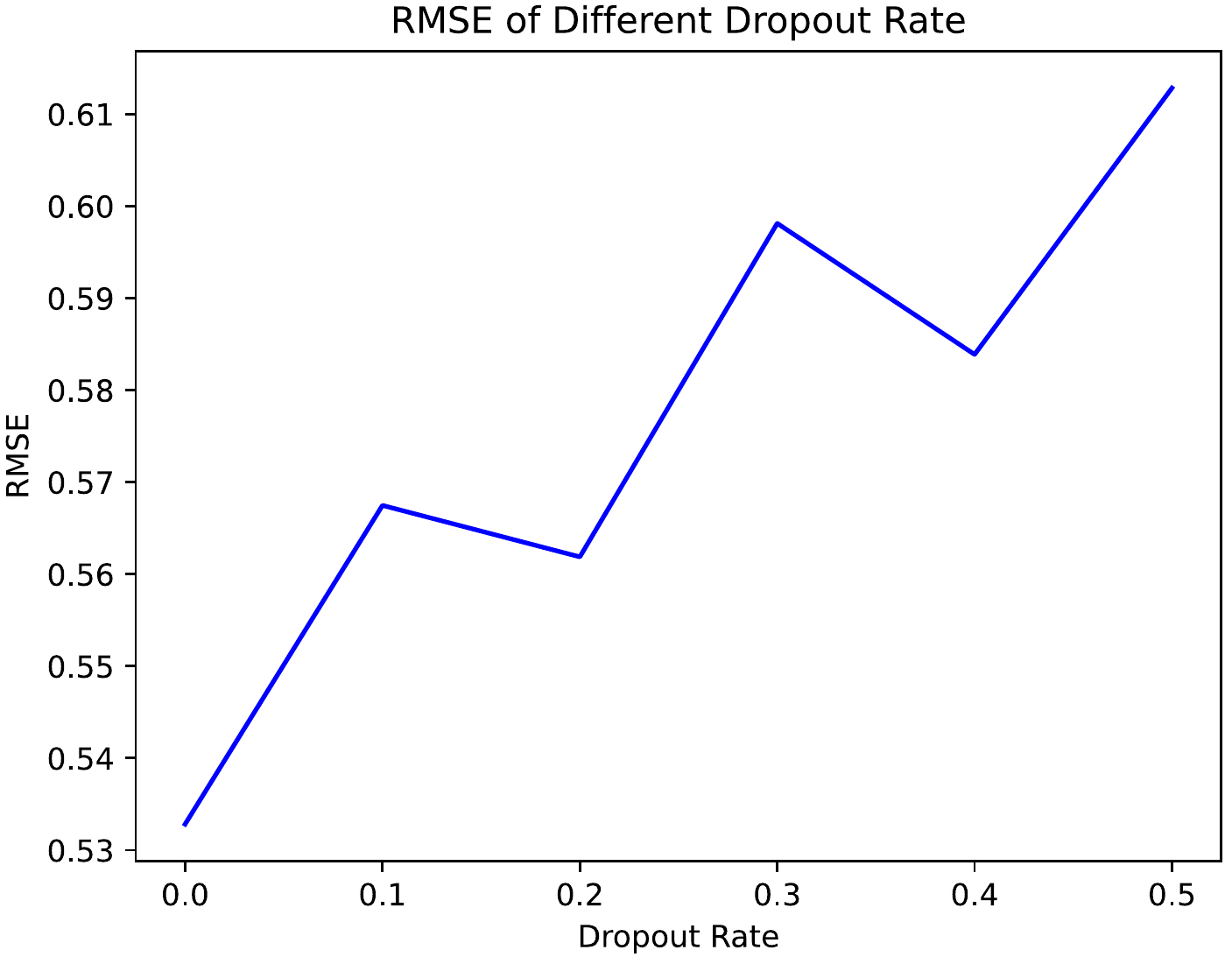}
    \caption{RMSE of Dropout Rate from 0.0 to 0.5 After 5,800 epochs}
    \label{fig:rmse}
\end{figure}

\begin{figure}
    \centering
    \includegraphics[width=0.5\linewidth]{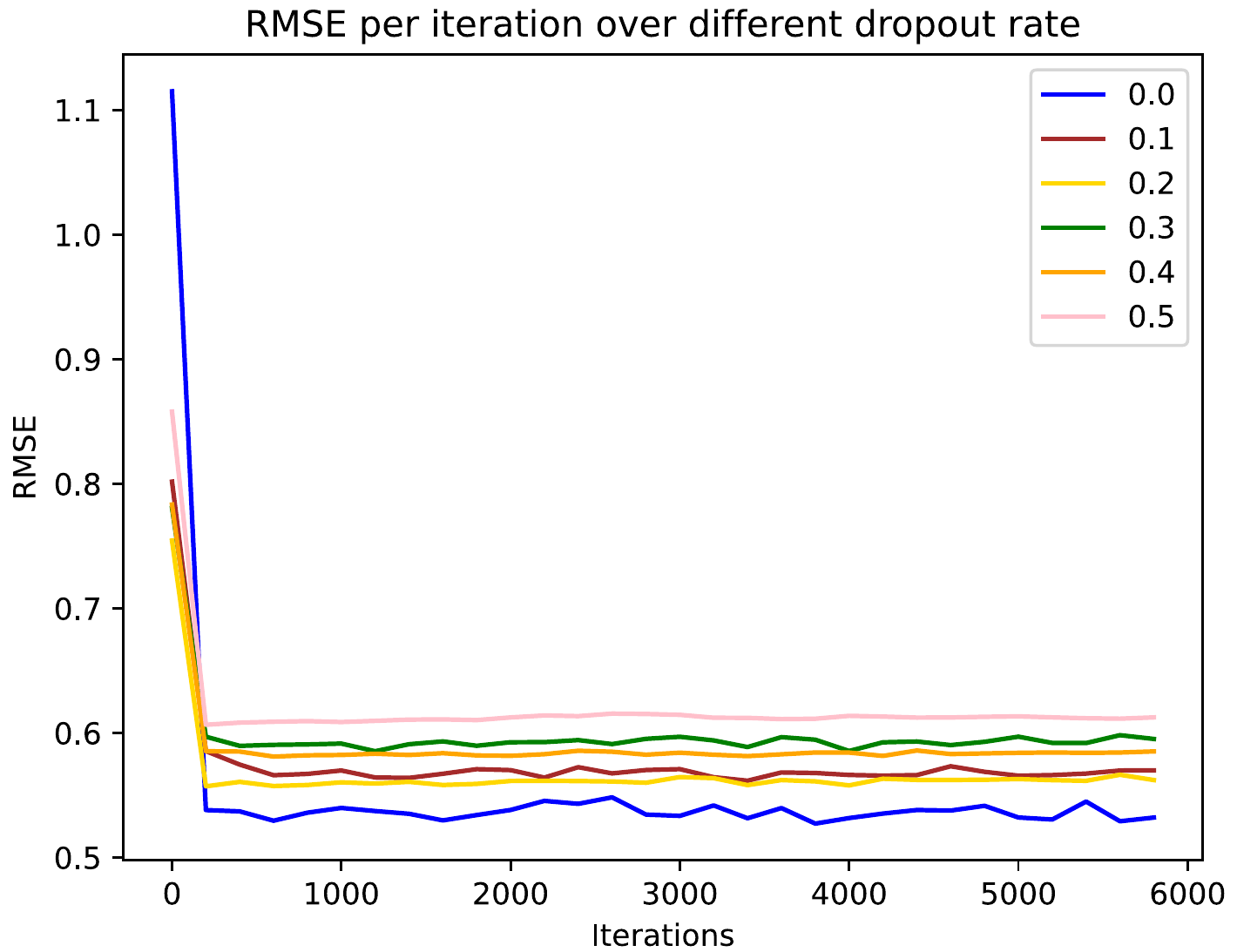}
    \caption{RMSE of Dropout Rate from 0.0 to 0.5 per Iteration}
    \label{fig:rmse_iter}
\end{figure}

Figure \ref{fig:rmse_iter} shows the convergence of different dropout rates per iteration. The figure illustrates that larger dropout rate converges to higher RMSE. It is clear that RMSE increased as the dropout rate increases in Figure \ref{fig:rmse}. Without any dropout, the RMSE is 0.533. When the dropout rate raises to 0.5, RMSE becomes 0.612. Dropout rate of 0.3 is a key setting. It varies differently from 0.3 to 0.5, from 0.598 to 0.584 and then to 0.612. Therefore, we believe that a setting of dropout rate from 0.3 to 0.5 would be appropriate.

\section{Discussion}

\subsection{Choosing an Appropriate Dropout Rate}
\begin{figure}
    \centering
    \includegraphics[width=0.5\linewidth]{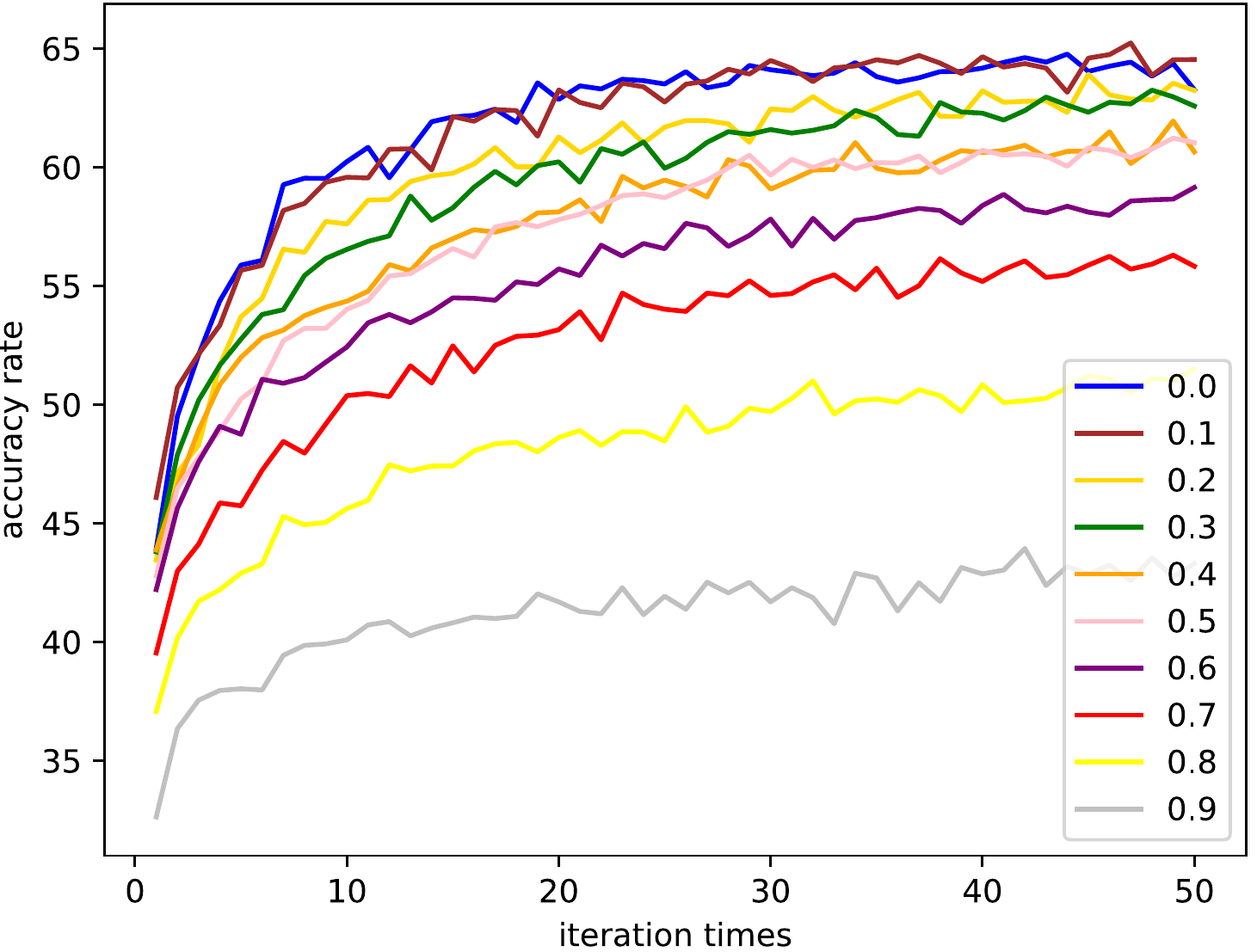}
    \caption{Classification Accuracy of LeNet-5 with Dropout Rate from 0.0 to 0.9}
    \label{fig:cifar_acc}
\end{figure}

In the former experiments Figure\ref{fig:0.0-0.5} and Figure\ref{fig:rmse}, it is not difficult to see that a dropout rate of 0.3-0.5 would be a perfect balance between model performance and accuracy. Increasing the dropout rate decreases the model performance (Figure \ref{fig:cifar_acc}), while decreasing the dropout rate makes the private data more likely to be revealed. Therefore, picking an appropriate dropout rate according to the application context is very important. For example, if the model trains on some sensitive user data, we can set the dropout rate to 0.5 to prevent attacks on the model. If the model trains on publicly available data, we believe a dropout rate between 0 and 0.3 would achieve the best performance.

\subsection{Trading Accuracy with Privacy}

Although the dropout layer is a very welcoming feature in natural language processing, neural networks in computer vision uses it much less often because of decrease in accuracy. However, sometimes it is a necessary evil trading accuracy with privacy, given the accuracy is still acceptable. In other previous work, such as DataMix and TextHide, all of them cannot avoid decrease in accuracy. 

\subsection{Dropout is NOT a Final Solution}

As you can from the experiment results, we can still recover some of the data of the original image even with a relatively high dropout rate. Therefore, we believe using cryptography methods remains to be the safest approach and maintains accuracy, if the training data demands a high level of security and the model demands high performance.

In \citet{zhao2020idlg}'s experiments, their algorithm is able to achieve 100\% accuracy in exacting labels. Sadly, adding an additional dropout layer cannot prevent that from happening.

\section{Conclusion}

In this paper, we propose using an additional dropout layer to prevent deep leakage attacks from recovering the local training data. As an undeniable threat to federal learning, Deep leakage is so powerful that we can only prevent it when we sacrifice some level of accuracy. While privacy has become most people's major concern nowadays, we hope that this work will inspire people to explore more methods in preserving private data security while maintaining a high level of performance of the original model.

\medskip

{
\small
\bibliographystyle{plainnat}
\bibliography{references}
}

\end{document}